\documentclass[11pt,a4paper]{article}
\usepackage[hyperref]{acl2020}
\usepackage{times}
\usepackage{latexsym}

\usepackage{microtype}

\usepackage{amsmath}
\usepackage{cleveref}
\usepackage{xcolor}
\usepackage{graphicx}
\usepackage{subfigure}
\usepackage{multirow}
\usepackage{booktabs}
\usepackage{comment}

\newcommand{\bfnew}{\\[5pt]}
\newcommand{\timemask}{ $\langle${\ttfamily TIME}$\rangle$ }
\newcommand{\moneymask}{ $\langle${\ttfamily MONEY}$\rangle$ }
\newcommand{\bymask}{ $\langle${\ttfamily BY-MASK}$\rangle$ }
\newcommand{\frommask}{ $\langle${\ttfamily FROM-MASK}$\rangle$ }

\newcommand{\unifootnotesize}{\footnotesize}
\newcommand{\blankspacecontrol}{\\[3pt]}

\aclfinalcopy 

\title{Measuring Forecasting Skill from Text}

\author{Shi Zong$^{1}$\quad Alan Ritter$^{1}$\quad Eduard Hovy$^{2}$ \\
$^{1}$Department of Computer Science and Engineering, The Ohio State University\\
$^{2}$Language Technologies Institute, Carnegie Mellon University\\
  \texttt{\{zong.56, ritter.1492\}@osu.edu, hovy@cs.cmu.edu}}

\date{}

\begin{document}
\maketitle
\begin{abstract}
  People vary in their ability to make accurate predictions about the future.  Prior studies have shown that some individuals can predict the outcome of future events with consistently better accuracy.  This leads to a natural question: what makes some forecasters better than others?  In this paper we explore connections between the language people use to describe their predictions and their forecasting skill.  Datasets from two different forecasting domains are explored: (1) geopolitical forecasts from Good Judgment Open, an online prediction forum and (2) a corpus of company earnings forecasts made by financial analysts.  We present a number of linguistic metrics which are computed over text associated with people's predictions about the future including: uncertainty, readability, and emotion.  By studying linguistic factors associated with predictions, we are able to shed some light on the approach taken by skilled forecasters.  Furthermore, we demonstrate that it is possible to accurately predict forecasting skill using a model that is based solely on language.  This could potentially be useful for identifying accurate predictions or potentially skilled forecasters earlier.\footnote{\unifootnotesize We provide our code and dataset descriptions at:\\ \url{https://github.com/viczong/measuring_forecasting_skill_from_text}.}
\end{abstract}


\section{Introduction}

People often make predictions about the future, for example meteorologists tell us what the weather might look like tomorrow, financial analysts predict which companies will report favorable earnings and intelligence analysts evaluate the likelihood of future geopolitical events.  An interesting question is why some individuals are significantly better forecasters \cite{mellers2015identifying}?

Previous work has analyzed to what degree various factors (intelligence, thinking style, knowledge of a specific topic, etc.) contribute to a person's skill.  These studies have used surveys or psychological tests to measure dispositional, situational and behavioral variables \cite{pub.1052640113}.  Another source of information has been largely overlooked, however: the language forecasters use to justify their predictions.  Recent research has demonstrated that it is possible to accurately forecast the outcome of future events by aggregating social media users' predictions and analyzing their veridicality \cite{swamy-etal-2017-feeling}, but to our knowledge, no prior work has investigated whether it might be possible to measure a forecaster's ability by analyzing their language.

In this paper, we present the first systematic study of the connection between language and forecasting ability. To do so, we analyze texts written by top forecasters (ranked by accuracy against ground truth) in two domains: geopolitical forecasts from an online prediction forum, and company earnings forecasts made by financial analysts.
To shed light on the differences in approach employed by skilled and unskilled forecasters, we investigate a variety of linguistic metrics.  These metrics are computed using natural language processing methods to analyze sentiment \citep{pang2002thumbs,wilson2005recognizing}, uncertainty \citep{de2012did,sauri2012you}, readability, etc.  In addition we make use of word lists taken from the Linguistic Inquiry and Word Count (LIWC) software \cite{doi:10.1177/0261927X09351676}, which is widely used in psychological research.  By analyzing forecasters' texts, we are able to provide evidence to support or refute hypotheses about factors that may influence forecasting skill.  For example, we show forecasters whose justifications contain a higher proportion of uncertain statements tend to make more accurate predictions.  This supports the hypothesis that more open-minded thinkers, who have a higher tolerance for ambiguity tend to make better predictions \cite{philip2005expert}.

Beyond analyzing linguistic factors associated with forecasting ability, we further demonstrate that it is possible to identify skilled forecasters and accurate predictions based only on relevant text. Estimating the quality of a prediction using the forecaster's language could potentially be very beneficial.  For example, this does not require access to historical predictions to evaluate past performance, so it could help to identify potentially skilled individuals sooner.  Also, forecasters do not always provide an explicit estimate of their confidence, so a confidence measure derived directly from text could be very useful.

\section{Linguistic Cues of Accurate Forecasting}
\label{sec:gjp_identify_skill}

In this section, we are interested in uncovering linguistic cues in people's writing that are predictive of forecasting skill.  We start by analyzing texts written by forecasters to justify their predictions in a geopolitical forecasting forum.
Linguistic differences between forecasters are explored by aggregating metrics across each forecaster's predictions.
In \S \ref{sec:cfra}, we analyze the accuracy of individual predictions using a dataset of financial analysts' forecasts towards companies' (continuous) earnings per share.  By controlling for differences between analysts and companies, we are able to analyze intra-analyst differences between accurate and inaccurate forecasts.

\subsection{Geopolitical Forecasting Data}
\label{sec:gjp_data}

To explore the connections between language and forecasting skill, we make use of data from Good Judgment Open,\footnote{\tiny \url{https://www.gjopen.com/}} an online prediction forum.  Users of this website share predictions in response to a number of pre-specified questions about future events with uncertain outcomes, such as: {\em ``Will North Korea fire another intercontinental ballistic missile before August 2019?''}  Users' predictions consist of an estimated chance the event will occur (for example, 5\%) in addition to an optional text justification that explains why the forecast was made.  A sample is presented in \Cref{fig:gjo_sample_ques}.

\begin{figure}[ht]
    \small
    \centering
    \begin{tabular}{|p{0.4 \textwidth}|}
        \hline
        {\bf Question:}  Will Kim Jong Un visit Seoul before 1 October 2019? \\\hline
        {\bf Estimated Chance:}  5\% \\\hline
        {\bf Forecast Justification:} No North Korean leader has stepped foot in Seoul since the partition of the Koreas at the end of the Korean War. \ldots\\
        \hline
    \end{tabular}
    \caption{A sample prediction made by a user in response to a question posted by \textit{the Economist}.}
    \label{fig:gjo_sample_ques}
\end{figure}

\vspace{0.1cm}
\noindent \textbf{Preprocessing.}
Not all predictions contain associated text justifications; in this work, we only consider predictions with justifications containing more than 10 tokens.
We ran {\ttfamily langid.py} \citep{lui-baldwin-2012-langid} to remove forecasts with non-English text, and further restrict our data to contain only users that made at least 5 predictions with text.

In our pilot studies, we also notice some forecasters directly quote text from outside resources (like Wikipedia, New York Times, etc.) as part of their justifications.  To avoid including justifications that are mostly copied from external sources, we remove forecasts that consist of more than 50\% text enclosed in quotation marks from the data. 
\blankspacecontrol
\textbf{Dataset statistics.}
We collected all questions with binary answers that closed before April 9, 2019, leading to a total of 441 questions.  23,530 forecasters made 426,909 predictions.
During preprocessing steps, 3,873 forecasts are identified as heavily quoted and thus removed.  After removing non-English and heavily quoted forecasts, forecasts with no text justifications or justifications less than 10 tokens, in addition users with fewer than 5 predictions with text, 55,099 forecasts made by 2,284 forecasters are selected for the final dataset.

The distribution of predictions made by each forecaster is heavily skewed. 8.0\% of forecasters make over 50 forecasts.\footnote{\unifootnotesize In our dataset, forecasters could even make over 1,000 forecasts with justifications.}  
On average, each forecaster makes 10.3 forecasts, excluding those who made over 50 predictions. In \Cref{tb:word_count}, we also provide breakdown statistics for top and bottom forecasters.

\subsection{Measuring Ground Truth}
\label{sec:gjp_methodology}
In order to build a model that can accurately classify good forecasters based on features of their language, we first need a metric to measure people's forecasting skill.  For this purpose we use Brier score \cite{brier1950verification}, a commonly used measure for evaluating probabilistic forecasts.\footnote{Other possible scoring rules exist, for example ranking forecasters by log-likelihood.
For a log-likelihood scoring rule, however, we need to adjust estimates of 1.00 and 0.00, which are not uncommon in the data, to avoid zero probability events.  There are many ways this adjustment could be done and it is difficult to justify one choice over another.}
For questions with binary answers, it is defined as:
\begin{align*}
\text{Forecaster's Brier Score} = \frac{1}{N}\sum_{i=1}^N (f_i - o_i)^2
\end{align*}
Here $f_i$ is the forecaster's estimated probability, $o_i$ is a binary variable indicating the final outcome of the event, and $N$ is the total number of forecasts.
Brier scores can be interpreted as the mean squared error between the forecast probability and true answer; lower scores indicate better forecasts.
\blankspacecontrol
\textbf{Ranking forecasters.} Directly comparing raw Brier scores is problematic, because users are free to choose
questions they prefer, and could achieve a lower Brier score simply by selecting easier questions.
To address this issue, we standardized Brier scores by subtracting the mean Brier scores and dividing by the standard deviation within questions \cite{pub.1052640113}.

We construct a set of balanced datasets for training and evaluating classifiers by choosing the top $K$ and bottom $K$ forecasters respectively.  In our experiments, we vary $K$ from 100 to 1,000; when $K$=1,000, the task can be interpreted roughly as classifying all $\sim$2k users into the top or bottom half of forecasters.\footnote{\unifootnotesize Readers may wonder if there do exist differences between top and bottom forecasters. We provide justifications for our ranking approach in \Cref{appendix:difference}.}

\subsection{Linguistic Analysis}
\label{sec:ling_analysis}

In \S \ref{sec:gjp_methodology}, we discussed how to measure ground-truth forecasting skill by comparing a user's predictions against ground-truth outcomes.
In the following subsections, we examine a selected series of linguistic phenomenon and their connections with forecasting ability.
Statistical tests are conducted using the paired bootstrap \citep{efron1994introduction}.  As we are performing multiple hypothesis testing, we also report results for Bonferroni-corrected significance level 0.05/30.

As discussed in \S \ref{sec:gjp_data}, the distribution of forecasts per user is highly skewed. 
To control for this, we compute averages for each forecaster and use aggregate statistics to compare differences between the two groups at the user-level. Analyses are performed over 6,639 justifications from the top 500 forecasters and 6,040 from bottom 500.

\subsubsection{Textual Factors}
\label{sec:textual_features}

\textbf{Length.} We first check the average length of justifications from different groups and report our results in \Cref{tb:word_count}. We observe that skilled forecasters normally write significantly longer justifications with more tokens per sentence. 
This suggests that good forecasters tend to provide more rationale to support their predictions.
\begin{table}[!h]
\centering
\resizebox{0.48\textwidth}{!}{%
\begin{tabular}{lccc}
\toprule
Metric & Top 500 & Btm 500 & $p$ \\\midrule
\textbf{Forecasters statistics}\\
\# users making $\geq$ 50 forecasts & 20 & 14 & - \\
Avg. forecasts (w/o above users) & 9.4 & 9.2 & -
\\[0.2cm]
\textbf{Length \& word counts} \\
Avg. \# tokens per user & 69.1 & 47.0 & $\uparrow\uparrow\uparrow$\\
\% answers $\geq$ 100 tokens per user & 18.5 & 8.3 & $\uparrow\uparrow\uparrow$ \\
Avg. \# tokens per sentence & 20.9 & 19.2 & $\uparrow\uparrow\uparrow$
\\
\hline
\end{tabular}
}
\caption{Statistics of our dataset. 
$p$-values are calculated by bootstrap test. $\uparrow\uparrow\uparrow$: $p$ $<$ 0.001.}
\label{tb:word_count}
\end{table}
\vspace{3pt}
\\
\noindent \textbf{Readability.} 
We compute two widely used metrics for readability: (1) Flesch reading ease \citep{Flesch:1948hk} and (2) Dale-Chall formula \citep{10.2307/1473169}. \Cref{tb:ling_difference} summarizes our results on average readability scores. We find good forecasters have lower readability compared to bad forecasters.

It is interesting to compare this result with the findings reported by \citet{ganjigunte-ashok-etal-2013-success}, who found a negative correlation between the success of novels and their readability, and also the work of \citet{sawyer2008readability} who found award winning articles in academic marketing journals had higher readability.  Our finding that more accurate forecasters write justifications that have lower readability suggests that skilled forecasters tend to use more complex language.

\begin{table}[!h]
\centering
\resizebox{0.45\textwidth}{!}{%
\begin{tabular}{llc}
\toprule
Metric & $p$ & Bonferroni\\\midrule
\multicolumn{3}{c}{\textit{Textual Factors}}
\\[0.05cm]
\textbf{Readability}\\
Flesch reading ease & $\downarrow\downarrow$ & \\
Dale-Chall & $\uparrow\uparrow\uparrow$ & $\ast$
\\[0.2cm]
\textbf{Emotion} \\
Absolute sentiment strength & $\downarrow\downarrow\downarrow$& $\ast$
\\[0.2cm]
\textbf{Parts of Speech}\\
Cardinal & $\uparrow\uparrow\uparrow$ & $\ast$\\
Noun & $\uparrow\uparrow$ \\
Preposition& $\uparrow\uparrow\uparrow$ & $\ast$\\
Pronoun & $\downarrow\downarrow\downarrow$ & $\ast$\\
\quad 1st personal pronoun & $\uparrow$\\
Verb & $\downarrow\downarrow\downarrow$ & $\ast$
\\
\midrule
\multicolumn{3}{c}{\textit{Cognitive Factors}}
\\[0.05cm]
\textbf{Uncertainty}\\
\% uncertain statements & $\uparrow\uparrow\uparrow$ & $\ast$\\
Tentative (LIWC) & $\uparrow\uparrow\uparrow$& $\ast$ 
\\[0.2cm]
\textbf{Thinking style}\quad\quad\quad\quad\\
\% forecasts with quoted text\quad\quad\quad & $\uparrow\uparrow\uparrow$ & $\ast$
\\[0.2cm]
\textbf{Temporal orientation}\\
Focus on past (LIWC) & $\uparrow\uparrow$ & \\
Focus on present \& future (LIWC) & $\downarrow\downarrow\downarrow$ & $\ast$\\
\bottomrule
\end{tabular}
}
\caption{\label{tb:ling_difference} Comparison of various metrics computed over text written by the top 500 and bottom 500 forecasters. Good forecasters tend to exhibit more uncertainty, cite outside resources, and tend toward neutral sentiment; they also use more complex language resulting in lower readability and focus more on past events. $p$-values are calculated by bootstrap test. The number of arrows indicates the level of $p$-value, while the direction shows the relative relationship between top and bottom forecasters, $\uparrow\uparrow\uparrow$: top group is higher than bottom group with $p$ $<$ 0.001, $\uparrow\uparrow$: $p$ $<$ 0.01, $\uparrow$: $p$ $<$ 0.05. Tests that pass Bonferroni correction are marked by $\ast$.}
\end{table}
\noindent \textbf{Emotion.} We also analyze the sentiment reflected in forecasters' written text. Rather than analyzing sentiment orientation (``positive'', ``negative'', or ``neutral''), here we focus on measuring sentiment \textit{strength}.
We hypothesize that skilled forecasters organize their supporting claims in a more rational way using less emotional language. 
Many existing sentiment analysis tools (e.g., \citet{socher-etal-2013-recursive}) are built on corpora such as the Stanford Sentiment Treebank, which are composed of movie reviews or similar texts. However, justifications in our dataset focus on expressing opinions towards future uncertain events, rather than simply expressing preferences toward a movie or restaurant, leading to a significant domain mismatch.  In pilot studies, we noticed many sentences that are marked as negative by the Stanford sentiment analyzer on our data do not in fact express a negative emotion. We thus use Semantic Orientation CALculator (SO-CAL), a lexicon-based model proposed by \citet{Taboada:2011:LMS:2000517.2000518} which has been demonstrated to have good performance across a variety of domains.
The model generates a score for each justification by adding together semantic scores of words present in the justification, with a 0 score indicating a neutral sentiment. We then take the absolute values of scores from the model and calculate averages for each group. Results in \Cref{tb:ling_difference} show that the top 500 forecasters have a significantly lower average sentiment strength compared to bottom 500 forecasters, indicating statements from skilled forecasters tend to express neutral sentiment.
\blankspacecontrol
\textbf{Parts of Speech.} 
As shown in Table \ref{tb:ling_difference}, we observe that top forecasters use a higher percentage of cardinal numbers and nouns, while higher numbers of verbs are associated with lower forecasting ability.\footnote{\unifootnotesize POS tags were obtained using Stanford CoreNLP. Nouns refer to common nouns.}

We also note the bottom 500 use a higher percentage of pronouns when justifying their predictions. To investigate this difference, we further separate first person pronouns\footnote{\unifootnotesize ``I'', ``me'', ``mine'', ``my'' and ``myself''.} from second or third person pronouns. As presented in \Cref{tb:ling_difference}, first person pronouns are used more often by the top forecasters.

\subsubsection{Cognitive Factors}
\label{sec:cognitive_features}

We now evaluate a number of factors that were found to be related to decision making processes based on prior psychological studies (e.g., \citet{pub.1052640113}), that can be tested using computational tools. A number of these metrics are calculated by using the Linguistic Inquiry and Word Count (LIWC) lexicon \citep{doi:10.1177/0261927X09351676}, a widely used tool for psychological and social science research.
\blankspacecontrol
\textbf{Uncertainty.}
To test the hypothesis that good forecasters have a greater tolerance for uncertainty and ambiguity, we employ several metrics to evaluate the degree of uncertainty reflected in their written language.
We use the model proposed by \citet{adel-schutze-2017-exploring} to estimate the proportion of uncertain statements made by each forecaster in our dataset.  It is an attention based convolutional neural network model, that achieves state-of-the-art results on a Wikipedia benchmark dataset from the 2010 CoNLL shared task \citep{farkas-etal-2010-conll}; we use the trained parameters provided by \citet{adel-schutze-2017-exploring}. 
After the model assigns an uncertainty label for each sentence, we calculate the percentage of sentences marked as uncertain.
Results of this analysis are reported in \Cref{tb:ling_difference}; we observe that the top 500 forecasters make a significantly greater number of uncertain statements compared to the bottom 500, supporting the hypothesis mentioned above.
\blankspacecontrol
\textbf{Thinking style.}
In \S \ref{sec:gjp_data}, we discussed the issue that many forecasts contain quoted text.  Although we removed posts consisting of mostly quoted text as a preprocessing step, we are interested in how people use outside resources during their decision making process.  We thus calculate the portion of forecasts with quotes for the two groups. We notice skilled forecasters cite outside resources more frequently. This may indicate that skilled forecasters tend to account for more information taken from external sources when making predictions.
\blankspacecontrol
\textbf{Temporal orientation.}
We make use of the LIWC lexicon \citep{doi:10.1177/0261927X09351676} to analyze the temporal orientation of forecasters' justifications.
We notice good forecasters tend to focus more on past events (reflected by tokens like \textit{``ago''} and \textit{``talked''}); bad forecasters pay more attention to what is currently happening or potential future events (using tokens like \textit{``now''}, \textit{``will''}, and \textit{``soon''}). We conjecture this is because past events can provide more reliable evidence for what is likely to happen in the future. 

\subsection{Predicting Forecasting Skill}
\label{sec:gjp_classification}

In \S \ref{sec:ling_analysis}, we showed there are significant linguistic differences between justifications written by skilled and unskilled forecasters.
This leads to a natural question: is it possible to automatically identify skilled forecasters based on the written text associated with their predictions?  We examine this question in general terms first, then present experiments using a realistic setup for early prediction of forecasting skill in \S \ref{sec:gjp_identify_early}.
\blankspacecontrol
\textbf{Models and features.} 
We start with a log-linear model using bag-of-ngram features extracted from the combined answers for each forecaster.  We experimented with different combinations of n-gram features from sizes 1 to 4.  N-grams of size 1 and 2 have best classification accuracy.
We map n-grams that occur only once to a $\langle${\ttfamily UNK}$\rangle$ token, and replace all digits with 0.
Inspired by our findings in \S \ref{sec:ling_analysis}, we also incorporate textual and cognition factors as features in our log-linear model.

We also experiment with convolutional neural networks \citep{kim-2014-convolutional} and BERT \citep{devlin-etal-2019-bert}. The 1D convolutional neural network consists of a convolution layer, a max-pooling layer, and a fully connected layer.  We minimize cross entropy loss using Adam \cite{iclr:adam}; the learning rate is 0.01 with a batch size of 32.  We fine-tune BERT on our dataset, using a batch size of 5 and a learning rate of 5e-6.  All hyperparameters were selected using a held-out dev set.
\blankspacecontrol
\textbf{Model performance.} 
Results are presented in \Cref{tb:gjo_classifier_performance}.  
As we increase the number of forecasters $K$, the task becomes more difficult as more forecasters are ranked in the middle.  However, we observe a stable accuracy around 70\%.  All models consistently outperform a random baseline (50\% accuracy), suggesting that the language users use to describe their predictions does indeed contain information that is predictive of forecasting ability. 
The n-grams with largest weights in the logistic regression model are presented in \Cref{tb_gjo:lr_top_ranked_features}.
We find that n-grams that seem to indicate uncertainty, including: {\em ``it seems unlikely''}, {\em ``seem to have''} and {\em ``it is likely''} are among the largest positive weights.

\begin{table}[!htbp]
\centering
\resizebox{0.46\textwidth}{!}{%
\begin{tabular}{cl|ccccc}
\toprule
& $K$ & 100 & 200 & 300 & 500 & 1000\\\midrule
\multirow{4}{*}{LR} & Bag-of-ngrams & 69.5 & 74.2 & 72.5 & 69.2 & 64.8\\
 & Textual & 66.0 & 60.8 & 62.0 & 59.3 & 57.4\\
& Cognitive & 69.0 & 68.0 & 67.3 & 65.5 & 61.0\\
& All above & 70.5 & 73.5 & 73.3 & 69.8 & 64.7 \\\midrule
\multirow{2}{*}{Neural} &  CNN & 71.5 & 75.0 & 72.0 & 69.6 & 64.0\\
& BERT-base & 74.5 & 77.3 & 74.3 & 69.7 & 65.1\\
\bottomrule
\end{tabular}
}
\caption{Accuracy (\%) on classifying skilled forecasters when choosing the top $K$ and bottom $K$ forecasters. For logistic regression (LR), we experiment with different sets of features: bag of \{1, 2\}-grams, textual factors in \S \ref{sec:textual_features}, cognitive factors in \S \ref{sec:cognitive_features}, and combination of all above.  For neural networks (Neural), we use convolutional neural network (CNN) and BERT-base. All results are based on 5-fold cross validation.}
\label{tb:gjo_classifier_performance}
\end{table}

\begin{table}[h!]
\centering
\resizebox{0.48\textwidth}{!}{%
\begin{tabular}{p{0.14\textwidth}|p{0.4\textwidth}}
\toprule
Top15\quad (High-weight)& in the next / . also , / . however , / based on the / there are no / . according to / of time . / . based on / they wo n't / there is no / it seems unlikely / do n't see / it is likely / more of a / seem to have 
\\\midrule
Bottom15\quad (Low-weight) & will continue to / it will be / the world . / . it 's / there is a / is not a / the west . / to be on / to be the / . yes , / he 's a / there will be / in the world / will still be / . he will
\\
\bottomrule
\end{tabular}
}
\caption{High and low-weight n-gram features from the logistic regression model trained to identify good forecasters ($K$=500 with only 3-gram features for interpretability). Positive features indicate some uncertainty (e.g., \textit{``it is likely''}, \textit{``seem to have''}
, \textit{``it seems unlikely''}), in addition to consideration of evidence from many sources (e.g., \textit{``based on the''}, \textit{``. according to''}).
}
\label{tb_gjo:lr_top_ranked_features}
\end{table}

\subsection{Identifying Good Forecasters Earlier}
\label{sec:gjp_identify_early}

With the model developed in \S \ref{sec:gjp_classification}, we are now ready to answer the following question: using only their first written justification, can we foresee a forecaster's future performance?
\bfnew \textbf{Setup.} Our goal is to rank forecasters by their performance.
We first equally split all 2,284 forecasters into two groups (top half versus bottom half) based on their standardized Brier scores. We then partition them into 60\% train, 20\% validation, and 20\% test splits within each group. We combine all justifications for each forecaster in the training set.  For forecasters in the validation and test sets, we only use their single earliest forecast.

We use forecasters' final rank sorted by averaged standardized Brier score over all forecasts as ground truth.  We then compare our text-based model to the following two baselines: (1) a random baseline (50\%) and (2) the standardized Brier score of the users' single earliest forecast.
\bfnew \textbf{Results.} 
We calculate the proportion of good forecasters identified in the top $N$, ranked by our text-based model, and report results in \Cref{tb:forecaster_first}. We observe that our models achieve comparable or even better performance relative to the first prediction's adjusted Brier score.  Calculating Brier scores requires knowing ground-truth, while our model can evaluate the performance of a forecaster \textit{without} waiting to know the outcome of a predicted event.

\begin{table}[!h]
\centering
\resizebox{0.45\textwidth}{!}{%
\begin{tabular}{l|ccc}
\toprule
 & P@10 & P@50 & P@100\\\midrule
Brier score & 60 & 64 & 62\\
\midrule
Text-based (LR) & 70 & 70 & 65\\
Text-based (CNN) & 90 & 68 & 64 \\
Text-based (BERT-base) & 80 & 70 & 67 \\
\bottomrule
\end{tabular}
}
\caption{\label{tb:forecaster_first} Precision@$N$ of identifying skilled forecasters based on their first prediction.}
\end{table}

\section{Companies' Earnings Forecasts}
\label{sec:cfra}

In \S \ref{sec:gjp_identify_skill}, we showed that linguistic differences exist between good and bad forecasters, and furthermore, these differences can be used to predict which forecasters will perform better.
We now turn to the question of whether it is possible to identify which {\em individual} forecasts, made by the same person, are more likely to be correct.  The Good Judgment Open data is not suitable to answer this question, because forecasts are discrete, and thus do not provide a way to rank individual predictions by accuracy beyond whether they are correct or not.  Therefore, in this section, we consider numerical forecasts in the financial domain, which can be ranked by their accuracy as measured against ground truth.

\noindent In this paper, we analyze forecasts of companies' earnings per share (EPS). Earnings per share is defined as the portion of a company's profit allocated to each share of common stock.  It is an important indicator of a company's ability to make profits.  
For our purposes, EPS also supports a cleaner experimental design as compared to stock prices, which constantly change in real time.
\bfnew \textbf{Data.} We analyze reports from the Center for Financial Research and Analysis (CFRA).\footnote{\tiny \url{https://www.cfraresearch.com/}} These reports provide frequent updates for analysts' estimates and are also organized in a structured way, enabling us to accurately extract numerical forecasts and corresponding text justifications.

We collected CFRA's analyst reports from the Thomson ONE database\footnote{\tiny \url{https://www.thomsonone.com/}} from 2014 to 2018. All notes making forecasts are extracted under the {\em ``Analyst Research Notes and other Company News''} section. The dataset contains a total of 32,807 notes from analysts, covering 1,320 companies.

\subsection{Measuring Ground Truth}
\label{sec:eps_classification}

We use a pattern-based approach (in \Cref{appendix:extract_numerical}) for extracting numerical forecasts. After removing notes without EPS estimates, 16,044 notes on 1,135 companies remain (this is after removing analysts who make fewer than 100 forecasts as discussed later in this section). We next evaluate whether the text can reflect how accurate these predictions are.
\blankspacecontrol
\textbf{Forecast error.} We measure the correctness of forecasts by absolute relative error \citep{BAREFIELD1975241, 10.2307/4479844}. The error is defined by the absolute difference between the analyst's estimate $e$ and corresponding actual EPS $o$, scaled by the actual EPS:
\begin{align*}
\text{Forecast Error} = \frac{|e-o|}{|o|}
\end{align*}
Low forecast errors indicate accurate forecasts.\footnote{Other methods for measuring the forecasting error have been proposed, for example to scale the relative error by stock price. We do not take this approach as stock prices are dynamically changing.}
\blankspacecontrol
\textbf{Ranking individual forecasts.}
As our goal is to study the intra-analyst differences between accurate and inaccurate forecasts, we standardize forecast errors within each analyst by subtracting the analyst's mean forecast error and then dividing by the standard deviation. To guarantee we have a good estimate for the mean, we only include analysts who make at least 100 forecasts (19 analysts are selected). 
We notice most forecast errors are smaller than 1, while a few forecasts are associated with very large forecasting errors.\footnote{\unifootnotesize For example, one analyst estimated an EPS for Fiscal Year 2015 of Olin Corporation (OLN) as \$1.63, while the actual EPS was \$-0.01, a standardized forecast error of 164.} Including these outliers would greatly affect our estimation for analysts' mean error. Thus, we only use the first 90\% of the sorted forecast errors in this calculation.

\subsection{Predicting Forecasting Error from Text}
\label{sec:classify_cfra_error}

Our goal is to test whether linguistic differences exist between accurate and inaccurate forecasts, independently of who made the prediction, or how difficult a specific company's earnings might be to predict.  To control for these factors, we standardize forecasting errors within analysts (as described in \S \ref{sec:eps_classification}), and create training/dev/test splits across companies and dates.
\blankspacecontrol
\textbf{Setting.} We collect the top $K$ and bottom $K$ predictions and split train, dev and test sets by time range and company. All company names are randomly split into 80\% train and 20\% evaluation sets. We use predictions for companies in the train group that were made in 2014-2016 as our training data. The dev set and test set consist of predictions for companies in evaluation group made during the years 2017 and 2018, respectively.  All hyperparameters are the same as those used in \S \ref{sec:gjp_classification}. When evaluating the classifier's performance, we balance the data for positive and negative categories.
\blankspacecontrol
\textbf{Results.} 
\Cref{tb:eps_classifier} shows the performance of our classifier on the test set. We observe our classifiers consistently achieve around 60\% accuracy when varying the number of top and bottom forecasts, $K$.  

\begin{table}[!htbp]
\centering
\resizebox{0.46\textwidth}{!}{%
\begin{tabular}{cl|cccc}
\toprule
& $K$ & 1000 & 2000 & 3000 & 5000 \\\midrule
\multirow{3}{*}{LR}& Bag-of-ngrams & 63.9 & 62.5 & 61.9 & 59.3 \\
& Linguistic & 56.3 & 59.2 & 55.4 & 55.5\\
& All above & 64.3 & 64.1 & 61.5 & 59.7\\\midrule
\multirow{2}{*}{Neural}& CNN & 66.7 & 67.8 & 64.7 & 64.0\\
& BERT-base & 70.8 & 66.7 & 65.8 & 64.4\\
\bottomrule
\end{tabular}
}
\caption{\label{tb:eps_classifier} Accuracy (\%) for classifying accurate predictions when using top $K$ and bottom $K$ analysts' predictions. We choose n-gram sizes to be 1 and 2. 
All reported results are on the test set.} 
\end{table}

\subsection{Linguistic Analysis}
\label{sec:ling_analysis_CFRA}

We present our linguistic analysis in \Cref{tb:ling_difference_cfra}. The same set of linguistic features in \S \ref{sec:ling_analysis} is applied to top 4,000 accurate and bottom 4,000 inaccurate analysts notes, excluding readability metric and quotation measure in thinking style metric.
Analysts' notes are written in a professional manner, which makes readability metric not applicable.  The notes do not contain many quoted text so we exclude quotation measure from the analysis.
We also replace the emotion metric with a sentiment lexicon specifically tailored for financial domain and provide our discussions. The Bonferroni-corrected significance level is 0.05/15.
We defer discussions to \S \ref{sec:domain_comparison} for comparing across different domains.
On average, each forecast contains 132.2 tokens with 5.5 sentences.
\blankspacecontrol
\noindent \textbf{Financial sentiment.} 
We make use of a lexicon developed by \citet{doi:10.1111/j.1540-6261.2010.01625.x}, which is specifically designed for financial domain. The ratio of positive and negative sentiment terms to total number of tokens is compared. Our results show that inaccurate forecasts use significantly more negative sentiment terms.

\begin{table}[!h]
\centering
\resizebox{0.45\textwidth}{!}{%
\begin{tabular}{llc}
\toprule
Metric & $p$ & Bonferroni \\\midrule
\textbf{Parts of Speech} &&\\
Cardinal & $\uparrow\uparrow$ & \\
Noun & $\uparrow\uparrow$ & \\
Verb & $\downarrow\downarrow\downarrow$ & $\ast$
\\[0.2cm]
\textbf{Uncertainty} & & \\
\% uncertain statements & $\downarrow\downarrow$ & $\ast$ 
\\[0.2cm]
\textbf{Temporal orientation} & & \\
Focus on past (LIWC) & $\uparrow\uparrow$ & $\ast$\\
Focus on present \& future (LIWC) \quad\quad& $\downarrow\downarrow\downarrow$ & $\ast$
\\[0.2cm]
\textbf{Financial sentiment} & & \\
Positive & $\uparrow\uparrow$ & \\
Negative & $\downarrow\downarrow\downarrow$ & $\ast$\\
\bottomrule
\end{tabular}
}
\caption{\label{tb:ling_difference_cfra} Comparison of various metrics over top 4,000 accurate and bottom 4,000 inaccurate forecasts. Only hypotheses with $p$ $<$ 0.05 are reported. See \S \ref{sec:ling_analysis_CFRA} for detailed justifications.
We follow the same notation as in \Cref{tb:ling_difference}, $\uparrow\uparrow\uparrow$: $p$ $<$ 0.001, $\uparrow\uparrow$: $p$ $<$ 0.01, $\uparrow$: $p$ $<$ 0.05.
}
\end{table}

\section{Comparison of Findings Across Domains}
\label{sec:domain_comparison}

In \S \ref{sec:gjp_identify_skill} and \S \ref{sec:cfra}, we analyze the language people use when they make forecasts in geopolitical and financial domains. Specifically, these two sections reveal how language is associated with accuracy both within and across forecasters. In this section, we compare our findings from these domains.

Our studies reveal several shared characteristics of accurate forecasts from a linguistic perspective over geopolitical and financial domains (in \Cref{tb:ling_difference} and \Cref{tb:ling_difference_cfra}). For example, we notice that skilled forecasters and accurate forecasts more frequently refer to past events. We also notice accurate predictions consistently use more nouns while unskilled forecasters use more verbs. 

We also note one main difference between two domains is uncertainty metric: in Good Judgment Open dataset, we observe that more skilled forecasters employ a higher level of uncertainty; while for individual forecasts, less uncertainty seems to be better. It makes us consider the following hypothesis: within each forecaster, people are more likely to be correct when they are more certain about their judgments, while in general skilled forecasters exhibit a higher level of uncertainty.
To test this hypothesis, we calculate the Spearman's $\rho$ between the financial analysts' mean forecasting errors and their average portion of uncertain statements.  Results show that these two variables are negative correlated with $\rho$=-0.24, which provides some support for our hypothesis, however the sample size is very small (there are only 19 analysts in the financial dataset). Also, these mean forecasting errors are not standardized by the difficulty of companies analysts are forecasting.

\section{Related Work}

Many recent studies have analyzed connections between users' language and human attributes \cite{hovy2015user,nguyen2013old,volkova2014inferring, Tan:2016:WAI:2872427.2883081, ICWSM148106}.
\citet{son2018causal} developed a tool for discourse analysis in social media and found that older individuals and females tend to use more causal explanations.
Another example is work by \citet{schwartz2015extracting}, who developed automatic classifiers for temporal orientation and found important differences relating to age, gender in addition to Big Five personality traits.
\citet{eichstaedt2015psychological} showed that language expressed on Twitter can be predictive of community-level psychological correlates, in addition to rates of heart disease.
\citet{graham-liu-2016-achieving} analyzed political polarization in social media and \citet{voigt2017language} examined the connections between police officers' politeness and race by analyzing language.
A number of studies \cite{DeChoudhury:2014:CPP:2531602.2531675,Eichstaedt11203,benton2017multitask,park2017living} have examined the connection between users' language on social media and depression and alcohol use \cite{using-longitudinal}. 
Other work has analyzed users' language to study the effect of attributes, such as gender, in online communication \citep{bamman2014gender,wang-jurgens-2018-going,voigt2018rtgender}.  In this work we study the relationship between people's language and their forecasting skill. To the best of our knowledge, this is the first work that presents a computational way of exploring this direction.

Our work is also closely related to prior research on predicting various phenomenon from users' language.  For example \citet{tan-etal-2014-effect} study the effect of wording on message propagation, \citet{gillick2018please} examine the connection between language used by politicians in campaign speeches and applause and \citet{perez2015experiments} explored linguistic differences between truthful and deceptive statements.  \citet{ganjigunte-ashok-etal-2013-success} show linguistic cues drawn from authors' language are strong indicators of the success of their books and \citet{tsur2009revrank} presented an unsupervised model to analyze the helpfulness of book reviews by analyzing their text.

There have been several studies using data from Good Judgment Open or Good Judgment Project \cite{mellers2015identifying}. 
One recent study examining the language side of this data is \citet{schwartz-etal-2017-assessing}.
Their main goal is to suggest objective metrics as alternatives for subjective ratings when evaluating the quality of recommendations. 
To achieve this, justifications written by one group are provided as tips to another group.  These justifications are then evaluated on their ability to persuade people to update their predictions, leading to \textit{real} benefits that can be measured by objective metrics. Prior work has also studied persuasive language on crowdfunding platforms \citep{yang2019let}.
In contrast, our work focuses on directly measuring forecasting skill based on text justifications. 

Finally we note that there is a long history of research on financial analysts' forecasting ability \citep{10.2307/245673, 10.2307/4480122, LOH2006455}. Most work relies on regression models to test if pre-identified factors are correlated with forecasting skill (e.g., \citet{LOH2006455, Call2009}). Some work has also explored the use of textual information in financial domain.  For example, \citet{kogan-etal-2009-predicting} present a study of predicting companies' risk by using financial reports. We also note a recent paper on studying financial analysts' decision making process by using text-based features from earning calls \cite{keith-stent-2019-modeling}. As far as we aware, our work is the first to evaluate analysts' forecasting skill based on their language.

\section{Limitations and Future Work}
Our experiments demonstrated it is possible to analyze language to estimate people's skill at making predictions about the future.  In this section we highlight several limitations of our study and ethical issues that should be considered before applying our predictive models in a real-world application. In our study, we only considered questions with binary answers; future work might explore questions with multiple-choice outcomes.
Prior studies have found that people's forecasting skills can be improved through experience and training \cite{doi:10.1177/0956797614524255}.  Our study does not take this into account as we do not have detailed information on the forecasters' prior experience.
Finally, we have not investigated the differences in our model's outputs on different demographic groups (e.g., men versus women), so our models may contain unknown biases and should not be used to make decisions that might affect people's careers.

\section{Conclusion}

In this work, we presented the first study of connections between people's forecasting skill and language used to justify their predictions.
We analyzed people's forecasts in two domains: geopolitical forecasts from an online prediction forum and a corpus of company earning forecasts made by financial analysts.
We investigated a number of linguistic metrics that are related to people's cognitive processes while making predictions, including: uncertainty, readability and emotion. 
Our experimental results support several findings from the psychology literature. For example, we observe that skilled forecasters are more open-minded and exhibit a higher level of uncertainty about future events.
We further demonstrated that it is possible to identify skilled forecasters and accurate predictions based solely on language.

\section*{Acknowledgments}
We would like to thank the anonymous reviewers for providing valuable feedback on an earlier draft of this paper.
This material is based in part on research sponsored by the NSF (IIS-1845670), ODNI and IARPA via the BETTER program (2019-19051600004) DARPA via the ARO (W911NF-17-C-0095) in addition to an Amazon Research Award. The views and conclusions contained herein are those of the authors and should not be interpreted as necessarily representing the official policies, either expressed or implied, of NSF, ODNI, ARO, IARPA, DARPA or the U.S. Government.

\bibliography{acl2020}
\bibliographystyle{acl_natbib}

\clearpage

\appendix

\section{Additional Experiments on Good Judgment Open Dataset}

\subsection{Differences Between Top and Bottom Forecasters?}
\label{appendix:difference}

\begin{figure}[h!]
\centering
\subfigure[Calibration curves by using all forecasts.]{
\begin{minipage}[t]{\linewidth}
\centering
\includegraphics[width=0.78\linewidth]{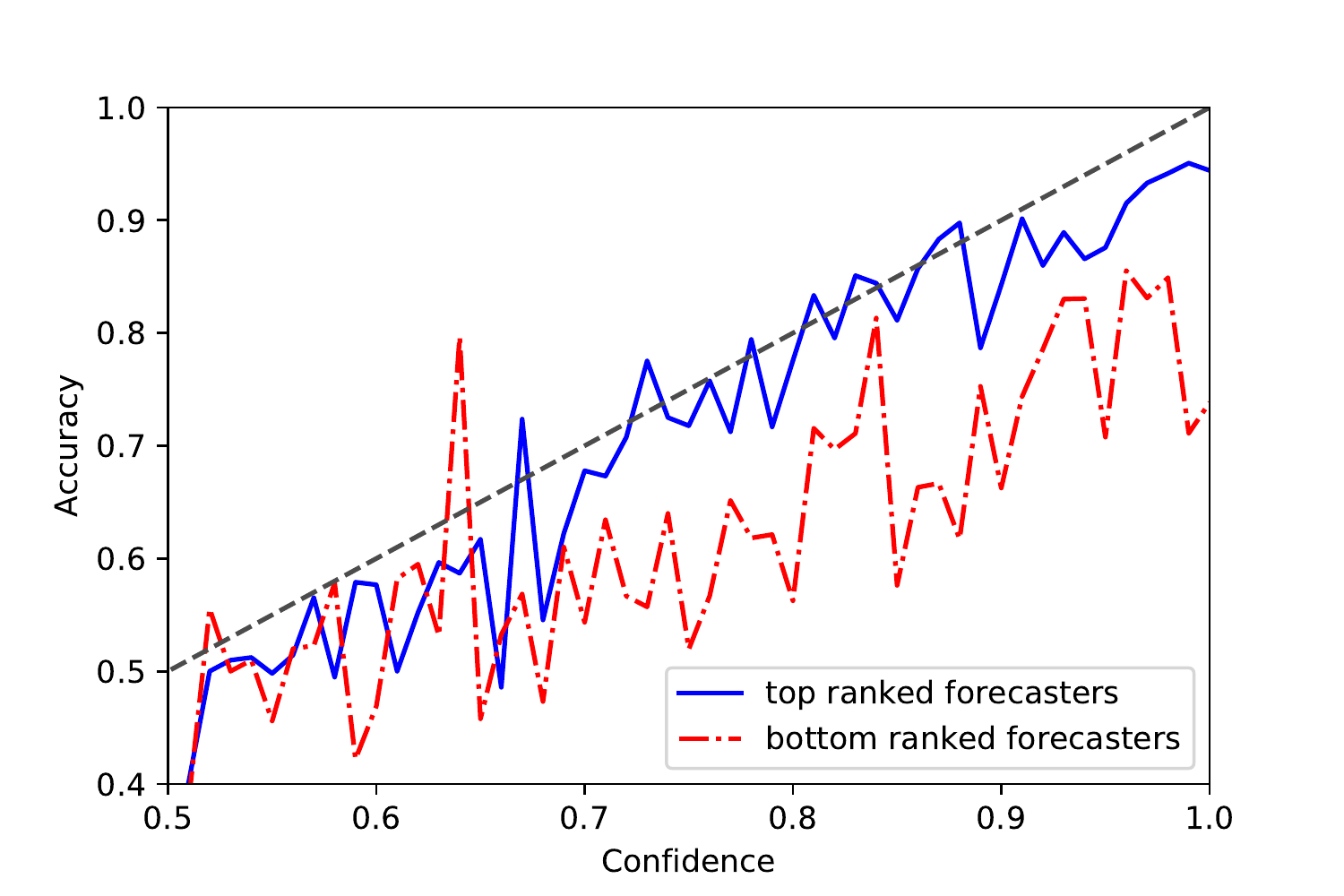}
\label{fig:existence_cali_all}
\end{minipage}
}
\subfigure[Aggregated forecasting performance across years.]{
\begin{minipage}[t]{\linewidth}
\centering
\includegraphics[width=0.78\linewidth]{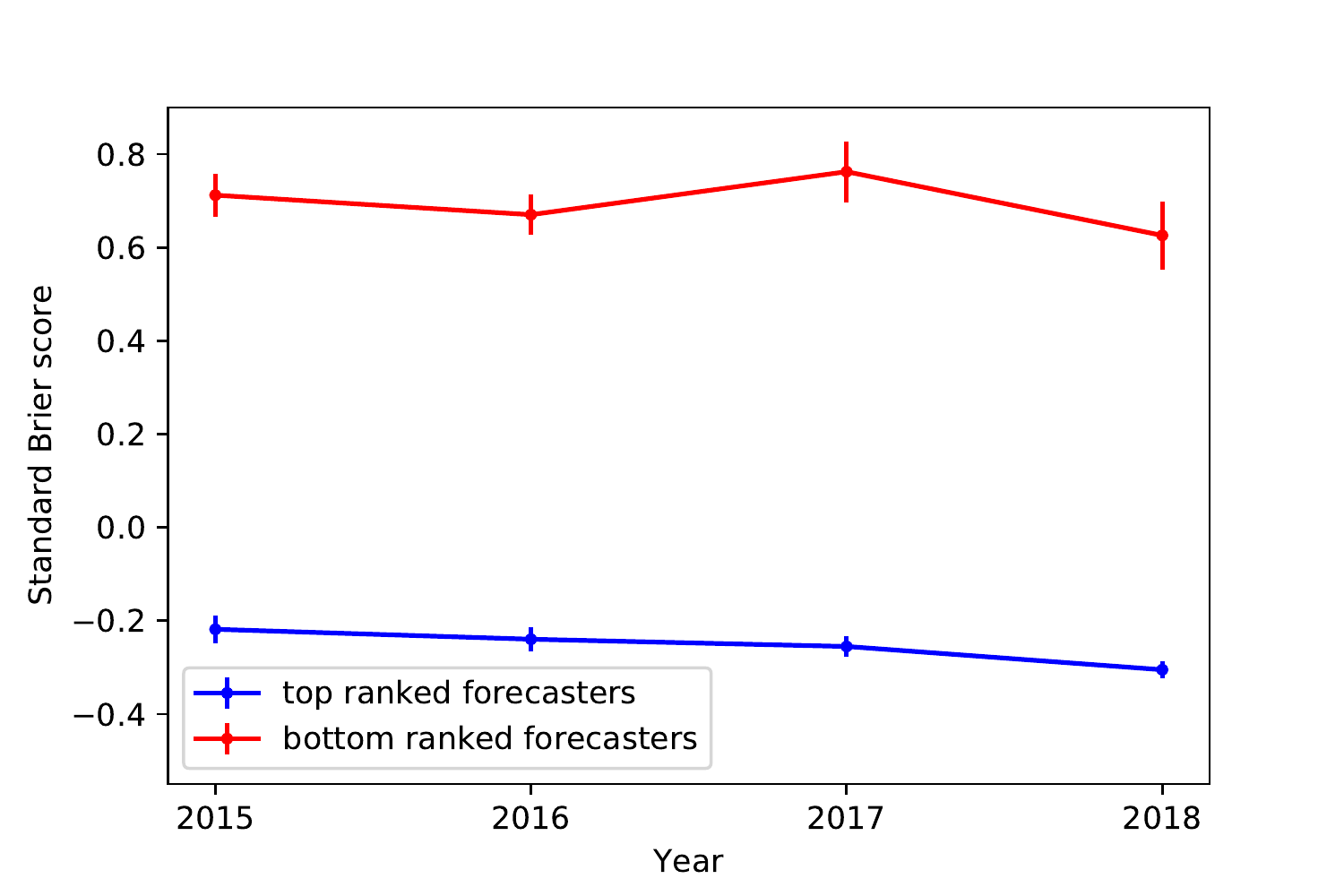}
\label{fig:existence_year}
\end{minipage}
}%
\caption{Comparison of forecasting skill between the top 500 and bottom 500 forecasters ranked by averaged standardized Brier scores. (a) Calibration curves for each group calculated using all forecasts (with and without justifications). The diagonal dotted line indicates a perfect calibration. (b) Trends of average standardized Brier scores over years. Negative values indicate better forecasting skill.}
\label{fig:cal}
\end{figure}

Figure \ref{fig:cal} presents calibration curves and averaged standardized Brier scores across years for the top and bottom 500 forecasters. We observe the differences between these two groups are persistent over time. Controlled lab experiments from psychology have also demonstrated that top forecasters ranked by Brier scores consistently have better forecasting performance than bottom forecasters \citep{pub.1052640113}.

\subsection{Additional Metrics and Examples for Linguistic Analysis}

\noindent \textbf{Uncertainty.} We present examples of sentences with uncertainty scores from our dataset in  \Cref{tb:uncertainty_sent_example}.

\noindent \textbf{Discourse connectives.}  
We further investigate the portion of discourse connectives used between sentences within each group. 
For this purpose, we use a lexicon developed by \citet{das-etal-2018-constructing}, which collects connectives from PDTB corpus connective list, RST Signalling Corpus and RST-DT relational indicator list. 
The lexicon contains 149 English connectives, divided into 4 categories: comparison, contingency, expansion, and temporal.\footnote{As some connectives are listed under more than one category, we restrict the list to those belonging to one or two categories.}
Our results show that skilled forecasters tend to use discourse connectives more frequently compared to unskilled forecasters, which may indicate that they tend to make more coherent arguments.
\blankspacecontrol
\textbf{Thinking style.} 
Analytical thinking score in LIWC \citep{doi:10.1177/0261927X09351676} ranks the level of a person's thinking skill. A high score correlates with formal, logical, and hierarchical thinking, while low scores are associated with informal, and narrative thinking. As shown in \Cref{tb:appendix_ling_difference}, good forecasters appear to demonstrate better analytical thinking skills.

\begin{table}[!h]
\centering
\resizebox{0.38\textwidth}{!}{%
\begin{tabular}{llc}
\toprule
Metric & $p$ & Bonferroni\\\midrule
\textbf{Discourse connectives}\\
Comparison & $\uparrow\uparrow\uparrow$ & $\ast$\\
Contingency & $\uparrow\uparrow$ & \\
Expansion & $\uparrow\uparrow$ & $\ast$\\
Temporal & $\uparrow\uparrow\uparrow$ & $\ast$
\\[0.2cm]
\textbf{Thinking style}\quad\quad\quad\quad\\
Analytical thinking (LIWC) & $\uparrow\uparrow$ & $\ast$\\
\bottomrule
\end{tabular}
}
\caption{\label{tb:appendix_ling_difference} Comparison of various metrics computed over text written by the top 500 and bottom 500 forecasters.
$p$-values are calculated by bootstrap hypothesis test. The number of arrows indicates the level of $p$-value, while the direction shows the relative relationship between top and bottom forecasters, $\uparrow\uparrow\uparrow$: top group is higher than bottom group with $p$ $<$ 0.001, $\uparrow\uparrow$: $p$ $<$ 0.01, $\uparrow$: $p$ $<$ 0.05. Tests that pass Bonferroni correction are marked by $\ast$.}
\end{table}

\begin{table*}[!h]
\centering
\resizebox{0.98\textwidth}{!}{%
\begin{tabular}{p{0.9\textwidth}|c}
\toprule
Sentence & Uncert. Score \\\midrule
Merkel is probably least prone to political scandals among the Western leaders and candidates . & 1.00\\\hline
It seems unlikely that the court would transfer the terms of that contract to Uber . & 0.99 \\\hline
My assumptions : - Sturgeon will not set a date for indyref2 before the UK elections on June 8 . & 0.05 \\\hline
To date , Toyota has distributed only 100 of the 300 Mirais preordered in California ... & 0.02 \\
\bottomrule
\end{tabular}
}
\caption{\label{tb:uncertainty_sent_example} Examples of sentences in our dataset with uncertainty scores estimated by the model proposed by \citet{adel-schutze-2017-exploring}. A higher uncertainty score indicates a higher level of uncertainty.}
\end{table*}

\begin{figure*}[h!]
\centering
\subfigure[Readability (Dale)]{
\begin{minipage}[h]{0.3\linewidth}
\centering
\includegraphics[width=0.99\linewidth]{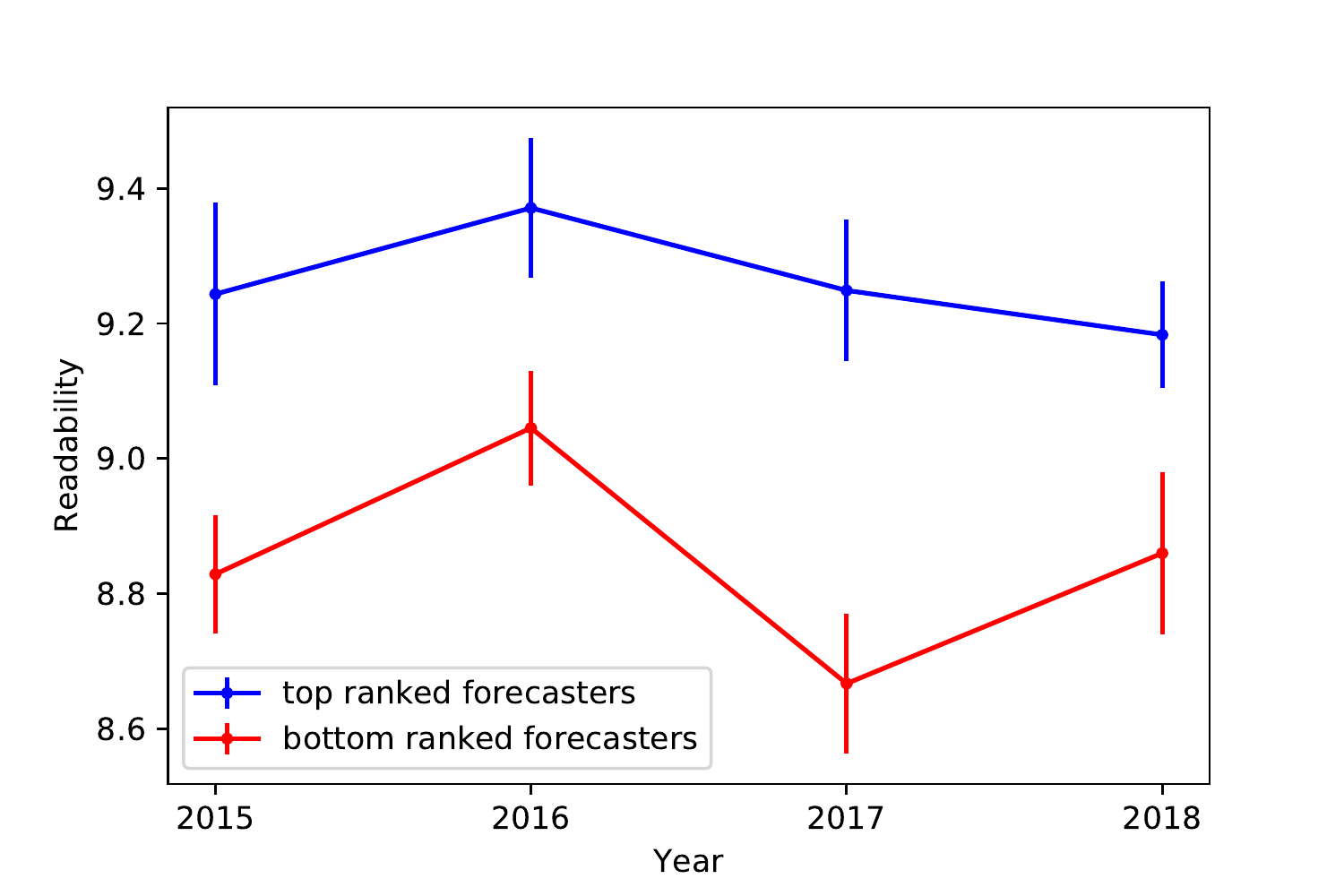}
\end{minipage}%
}%
\subfigure[Emotion]{
\begin{minipage}[h]{0.3\linewidth}
\centering
\includegraphics[width=0.99\linewidth]{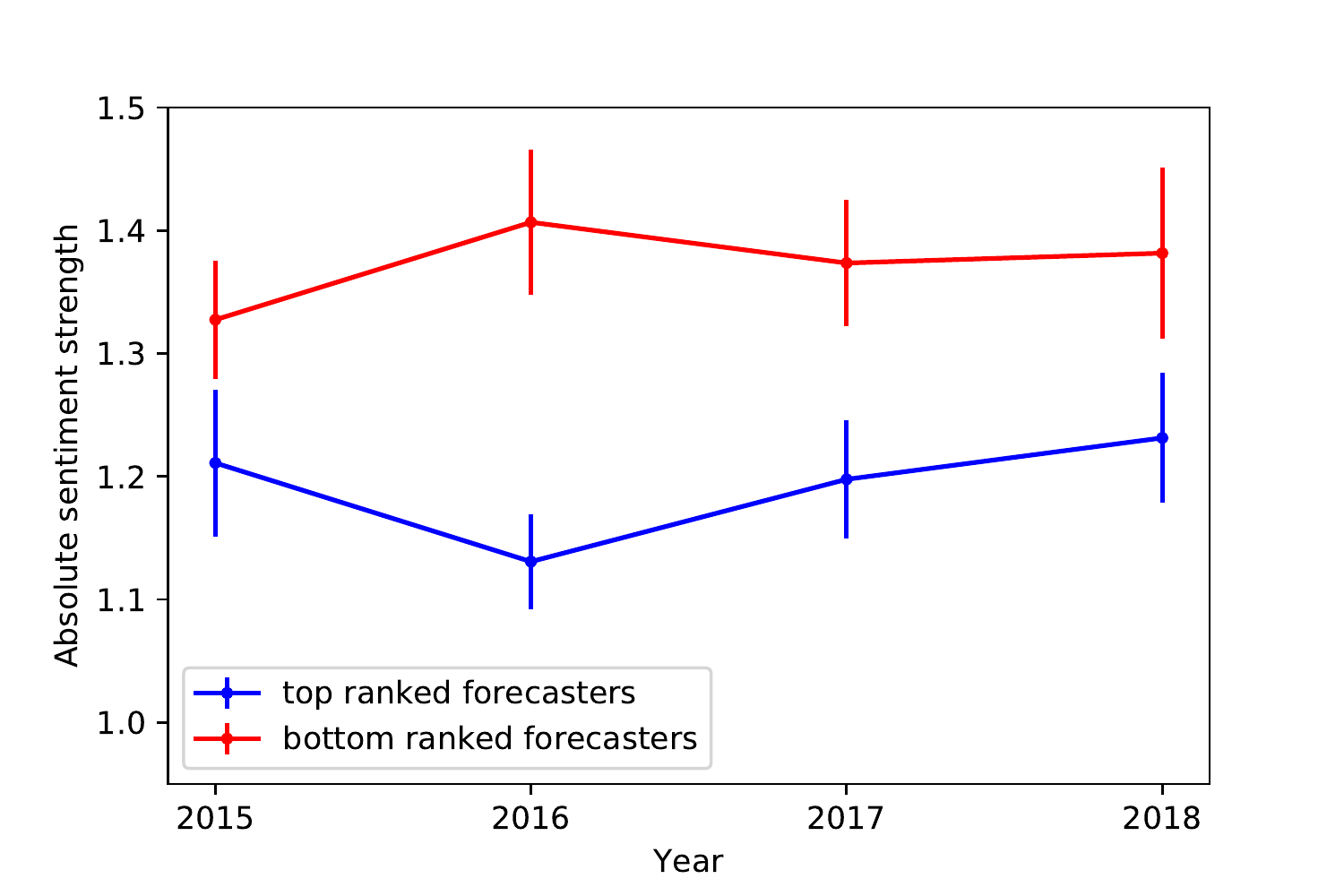}
\end{minipage}
}
\subfigure[Parts of Speech (noun)]{
\begin{minipage}[h]{0.3\linewidth}
\centering
\includegraphics[width=0.99\linewidth]{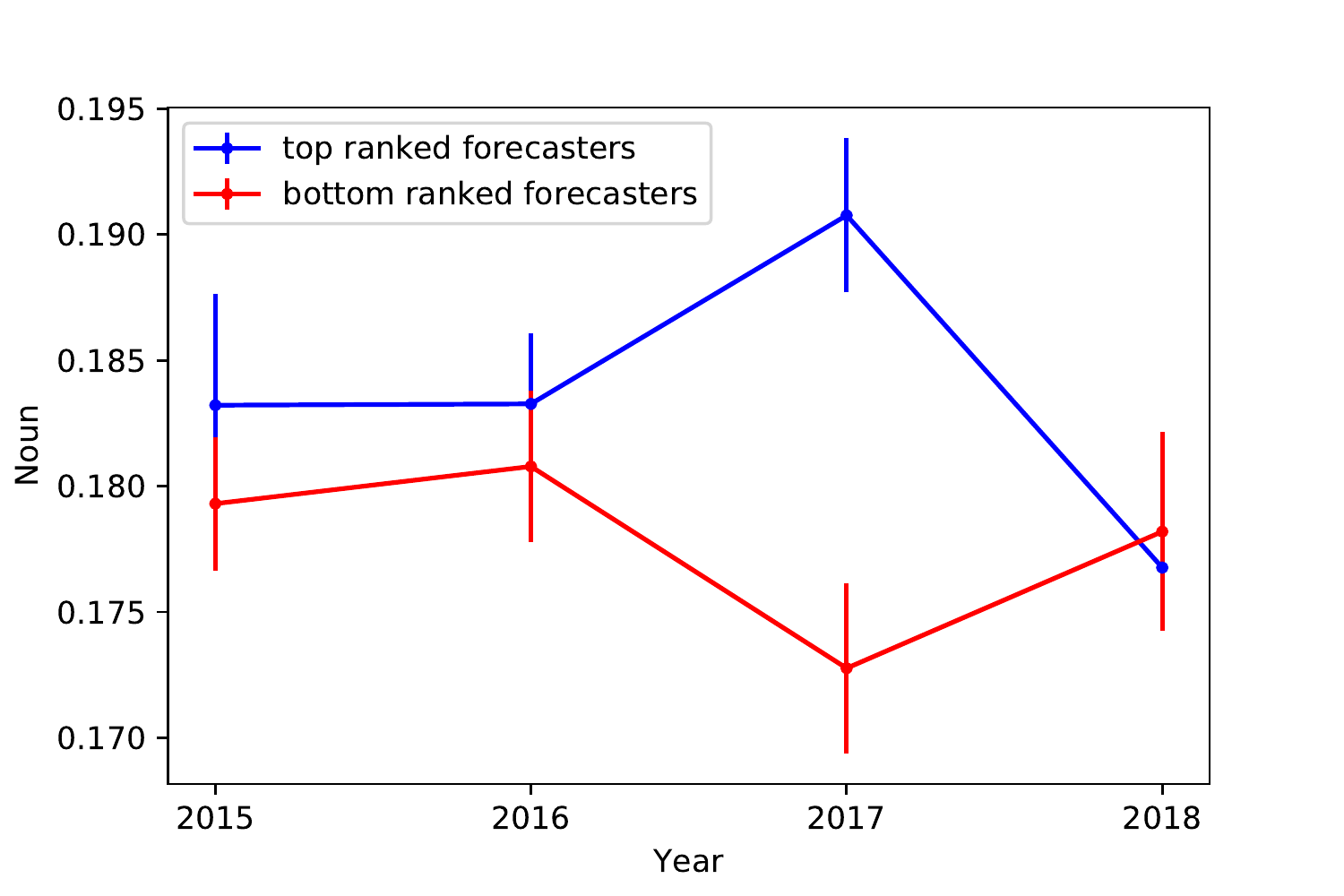}
\end{minipage}
}%
\\
\subfigure[Parts of Speech (verb)]{
\begin{minipage}[h]{0.3\linewidth}
\centering
\includegraphics[width=0.99\linewidth]{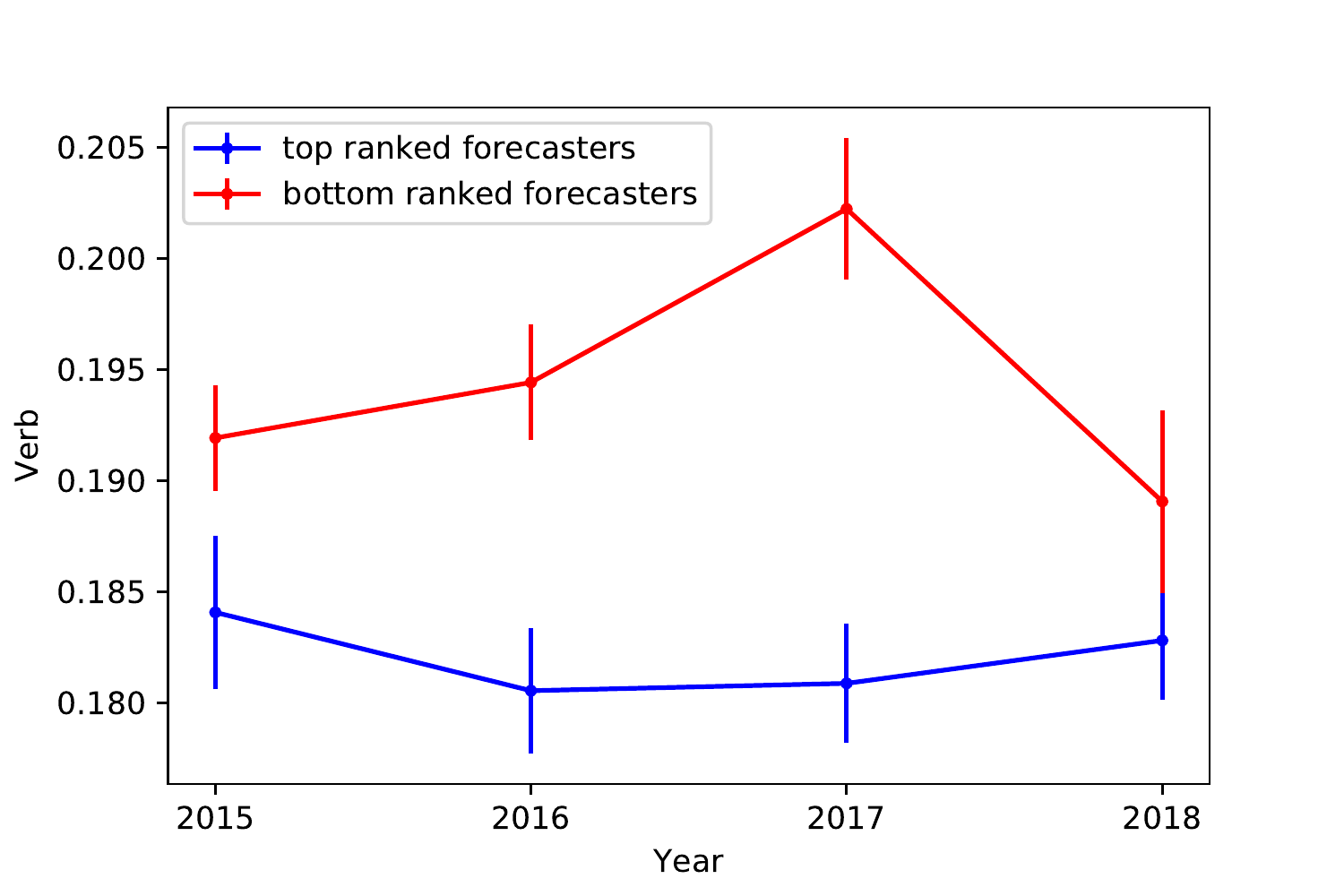}
\end{minipage}
}%
\subfigure[Discourse connectives (comparison)]{
\begin{minipage}[h]{0.3\linewidth}
\centering
\includegraphics[width=0.99\linewidth]{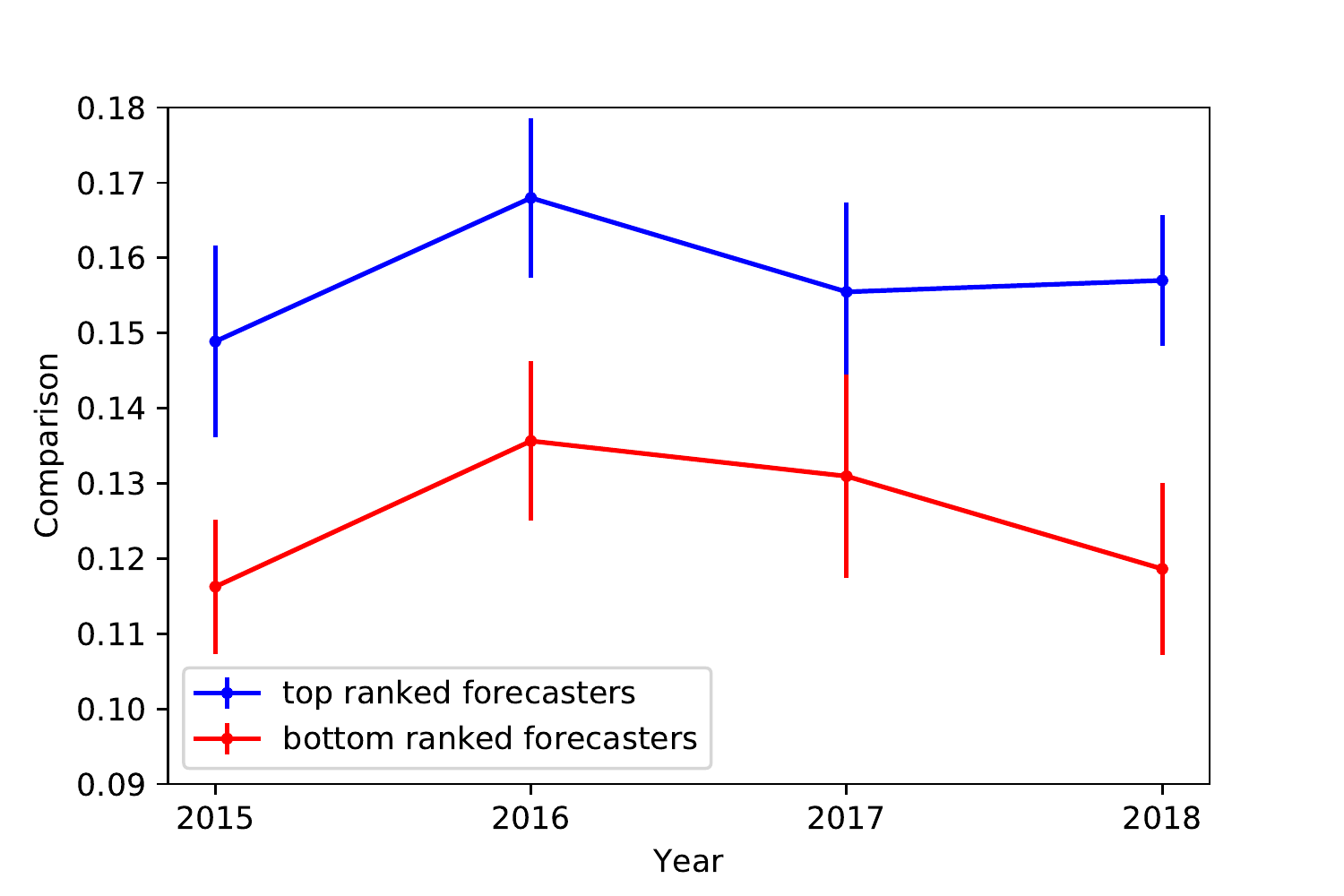}
\end{minipage}%
}%
\subfigure[Discourse connectives (temporal)]{
\begin{minipage}[h]{0.3\linewidth}
\centering
\includegraphics[width=0.99\linewidth]{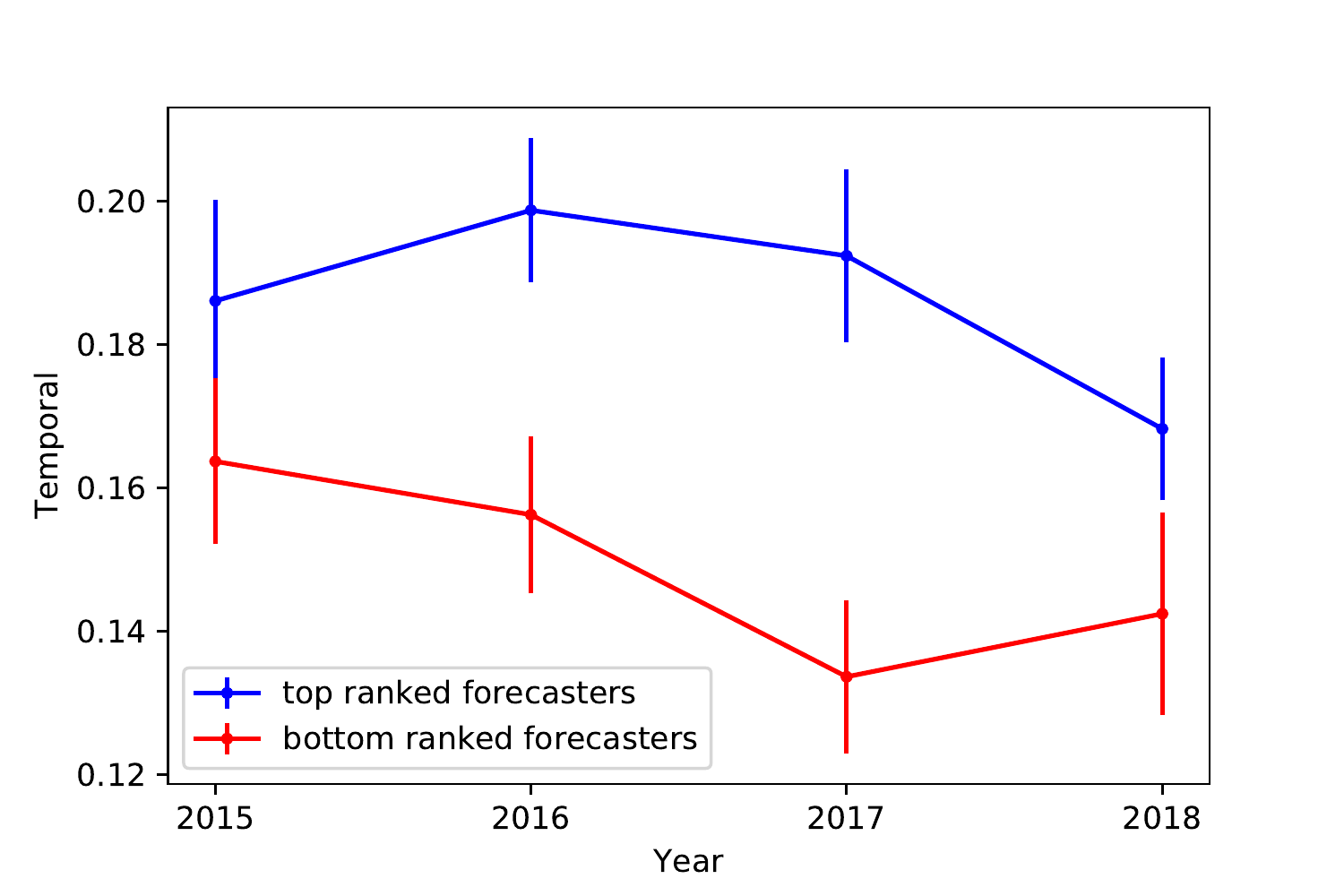}
\end{minipage}
}%
\\
\subfigure[Uncertainty]{
\begin{minipage}[h]{0.3\linewidth}
\centering
\includegraphics[width=0.99\linewidth]{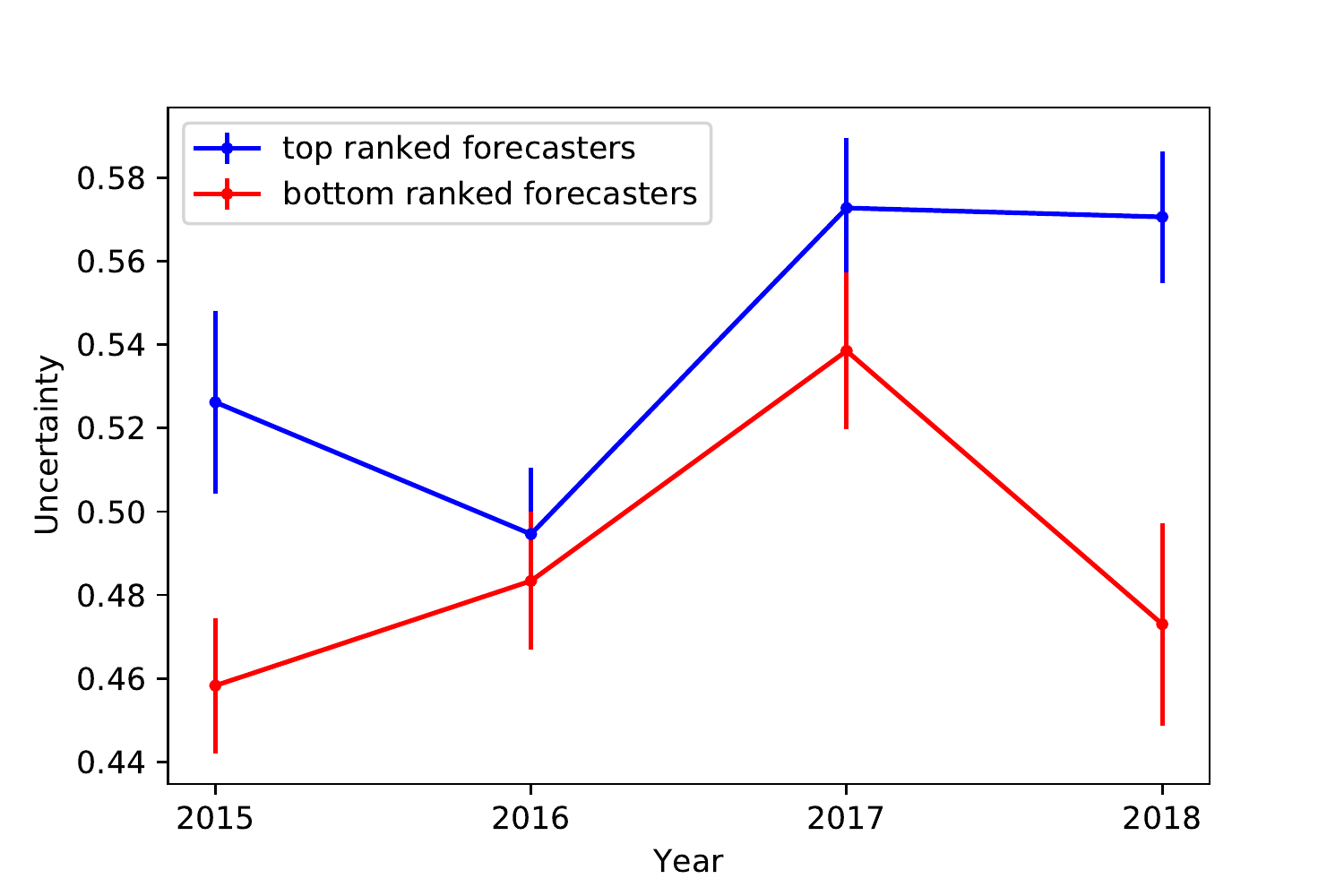}
\end{minipage}%
}%
\subfigure[Thinking style (analytical score)]{
\begin{minipage}[h]{0.3\linewidth}
\centering
\includegraphics[width=0.99\linewidth]{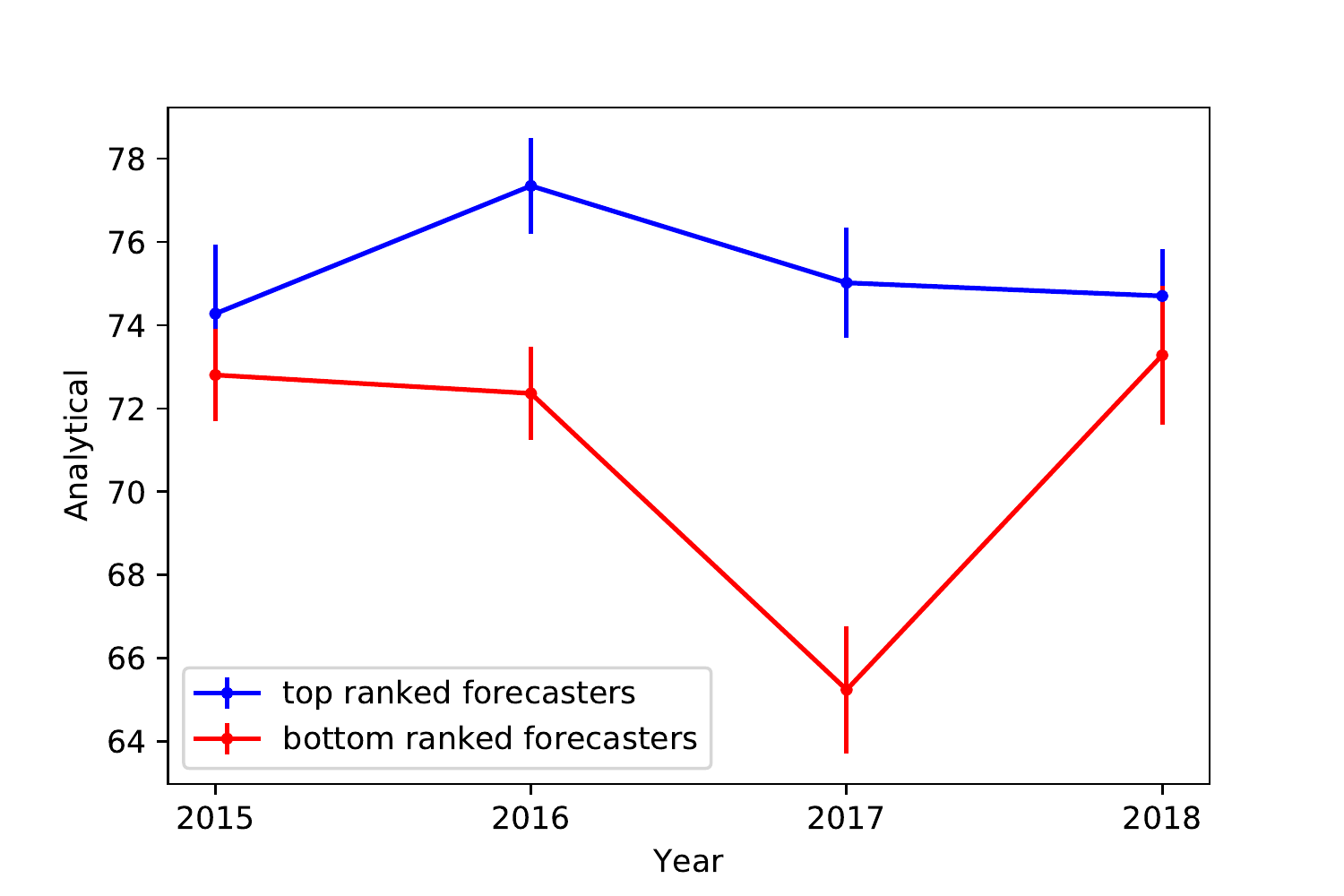}
\end{minipage}
}%
\subfigure[Temporal orientation (focus on past)]{
\begin{minipage}[h]{0.3\linewidth}
\centering
\includegraphics[width=0.99\linewidth]{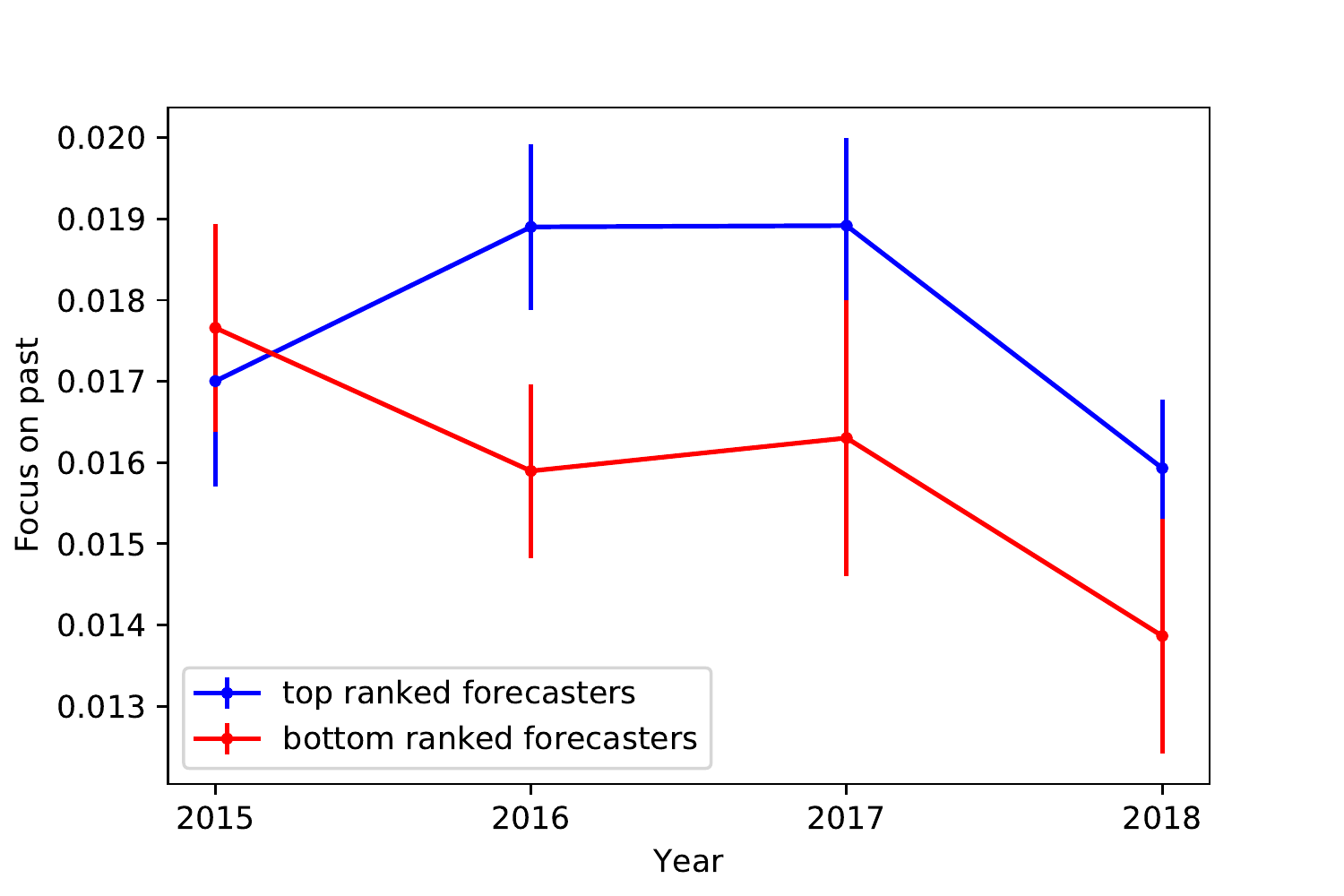}
\end{minipage}
}
\caption{Linguistic features in different years for top 500 and bottom 500 forecasters. The plots show how readability (Dale), emotion, Parts of Speech (noun and verb), discourse connectives (comparison and temporal), uncertainty, thinking style (analytical score), and temporal orientation (focus on past) change in different years.  We observe nearly consistent trends for all metrics over time, which indicates that linguistic differences are stable.  Error bars represent standard errors.}
\label{fig:ling_over_year}
\end{figure*}

\subsection{Linguistic Cues over Time} 

We are interested in whether our observed linguistic differences are consistent over time. To answer this question, we select the top 500 and bottom 500 forecasters based on their final ranking and evaluate aggregated metrics for the two groups in different years. Our results are shown in \Cref{fig:ling_over_year}. We observe the same pattern for all linguistic metrics. For example, skilled forecasters consistently exhibit a higher level of uncertainty and past temporal orientation, and a lower readability compared to unskilled forecasters.

\begin{table*}[!htbp]
\centering
\small
\resizebox{0.98\textwidth}{!}{%
\begin{tabular}{l|p{0.8\textwidth}}
\toprule
Sentence & We trim our 12-month target price to \$20 from \$23 , 10X our '16 EPS estimate of \$2.01  -LRB- trimmed today from \$2.10 -RRB- .\\
Pattern & \timemask EPS estimate of \moneymask \\
Extracted & $\langle$'16, \$2.01$\rangle$\\\midrule
Sentence & We raise '18 and '19 EPS estimates by \$4.61 and \$5.72 to \$19.85 and \$25.95 . \\
Pattern & \timemask and \timemask EPS estimates \bymask to \moneymask and \moneymask \\
Extracted & $\langle$'18, \$19.85$\rangle$, $\langle$'19, \$25.95$ \rangle$\\\midrule
Sentence & We raise our FY 17 EPS estimate to \$3.23 from \$2.96 and set FY 18 's at \$3.43 . \\
Pattern & \timemask EPS estimate to \moneymask \frommask and set \timemask at \moneymask\\
Extracted & $\langle$FY 17, \$3.23$\rangle$, $\langle$FY 18, \$3.43$\rangle$ \\ 
\bottomrule
\end{tabular}
}
\caption{\label{tb:eps_extraction_example} Examples of earnings forecasts extracted from analysts' notes.  Only sentences mentioning the earnings forecast are shown; the notes also contain additional analysis to justify the forecast. All sentences from notes are used to classify accurate versus inaccurate forecasts as described in \S \ref{sec:classify_cfra_error}.
}
\end{table*}

\section{Experimental Details on Companies' Earning Forecasts}
\subsection{Extracting Numerical Forecasts from Text}
\label{appendix:extract_numerical}

Not all analysts' notes in our dataset are associated with structured earnings forecasts (in tables).
Instead, the analysts' numerical predictions for future earnings are directly reported in the text of their notes, which also contain additional language justifying their predictions.  Therefore, our first goal is to extract structured representations of analysts' EPS estimates in a $\langle${\ttfamily TIME}, {\ttfamily VALUE}$\rangle$ format.  We noticed that analysts have a highly consistent style when writing this section of the report, we therefore use a set of lexico-syntactic patterns to extract the forecasts from text; as described below.  We found this approach to have both high precision and high recall.

We randomly sampled 60\% of the notes in our dataset for developing patterns.  Before generating the rules, we replaced entities indicating time and money with special \timemask and \moneymask tokens. 
To evaluate the generalization of our patterns, we randomly sampled 100 sentences containing 136 numerical forecasts from the remaining 40\% of notes and manually checked all of them.  We estimate that our pattern-based approach extracts numerical forecasts with 0.91 precision and 0.82 recall. \Cref{tb:eps_extraction_example} shows examples of numerical forecasts extracted using our approach.  In a few cases we found that an analyst's note can contain more than one forecast.  For simplicity, we only consider the earliest forecast that is made within the 2014-2018 time range.

\end{document}